\documentclass[runningheads]{llncs}
\usepackage[T1]{fontenc}
\usepackage{graphicx}
\usepackage{hyperref}
\usepackage[russian,british]{babel}
\begin{document}
\title{Key Algorithms for Keyphrase Generation: Instruction-Based LLMs for Russian Scientific Keyphrases}
\titlerunning{Key Algorithms for Keyphrase Generation}
%

\author{Anna Glazkova\inst{1}\orcidID{0000-0001-8409-6457} \and
Dmitry Morozov\inst{2}\orcidID{0000-0003-4464-1355} \and
Timur Garipov\inst{2}\orcidID{0009-0008-4527-2268}}
%
\authorrunning{A. Glazkova et al.}
%
\institute{
University of Tyumen, Tyumen, Russia\\
\email{a.v.glazkova@utmn.ru}\\\and
Novosibirsk State University, Novosibirsk, Russia\\
\email{morozowdm@gmail.com}, \email{garipov154@yandex.ru}}

\maketitle              
\begin{abstract}

Keyphrase selection is a challenging task in natural language processing that has a wide range of applications. Adapting existing supervised and unsupervised solutions for the Russian language faces several limitations due to the rich morphology of Russian and the limited number of training datasets available. Recent studies conducted on English texts show that large language models (LLMs) successfully address the task of generating keyphrases. LLMs allow achieving impressive results without task-specific fine-tuning, using text prompts instead. In this work, we access the performance of prompt-based methods for generating keyphrases for Russian scientific abstracts. First, we compare the performance of zero-shot and few-shot prompt-based methods, fine-tuned models, and unsupervised methods. Then we assess strategies for selecting keyphrase examples in a few-shot setting. We present the outcomes of human evaluation of the generated keyphrases and analyze the strengths and weaknesses of the models through expert assessment. Our results suggest that prompt-based methods can outperform common baselines even using simple text prompts.

\keywords{Keyphrase selection  \and Large language model \and Prompt-based learning \and Scientific texts.}
\end{abstract}
\section{Introduction}

A keyphrase or a keyword is a brief and summative content that captures the main idea of a longer text. In this work, the term ''keyphrase'' is used to refer to both keyphrases and keywords, and it means that a keyphrase can consist of one or more words. Effective keyphrases can enhance comprehension, organization, and retrieval of document content. They are widely used in digital libraries and searchable document collections for systematizing texts. In particular, a list of keyphrases is an important component of a research paper as they help summarize the main ideas discussed in a text, thus simplifying the process of information retrieval and selecting relevant papers. Current studies divide keyphrases into two categories: present and absent, i.e., those that are found or not found in the original text in explicit form~\cite{muhammad2024pre}. Keyphrase selection involves identifying and extracting significant phrases within the text and generating keyphrases that generalize or expand the text content. Keyphrase selection presents a complex challenge that requires the ability to summarize and comprehend the source text at a high-quality level. Recently, tasks related to deep text understanding have often been addressed using large language models~(LLMs).

In the past few years, LLMs have transformed the field of natural language processing (NLP) through their outstanding performance on various tasks. In particular, it has been found that LLMs can achieve high results without fine-tuning for specific tasks but only by applying text prompts containing necessary instructions \cite{dang2022prompt,hadi2023large}. Some recent studies \cite{martinez2023chatgpt,song2023large,song2023chatgpt} have applied prompt-based LLMs to generate keyphrases on English-language datasets and obtained competitive results compared to those of fine-tuned language models and unsupervised methods. The advantages of prompt-based models include the absence of the need to construct a large training dataset and the ability to generate both present and absent keyphrases in a lemmatized form.

Existing solutions for keyphrase selection in the Russian language mainly represent the adaptation of existing unsupervised methods. The possibilities of these methods are limited in view of the rich morphology of the Russian language~\cite{morozov2023generation,sheremetyeva2015}. Besides, the number of Russian datasets for keyphrase selection is limited. In this work, we explore the ability of open-source instruction-following LLMs to generate keyphrases for Russian scientific texts and compare their results with other common keyphrase selection solutions. The findings illustrate that prompt-based keyphrase generation can exhibit superior performance in comparison with current solutions for Russian. The contribution of this work is as follows: (i) we compare prompt-based LLMs for keyphrase generation with fine-tuned models and unsupervised methods; (ii) we evaluate zero-shot and few-shot prompting for keyphrase generation and study strategies for selecting keyphrase examples in a few-shot setting; (iii) we provide the results of human evaluation of the selected keyphrases and identify advantages and disadvantages of the models based on expert assessment. 

The paper is organized as follows. Section \ref{sec2} provides a brief review of related work. Section \ref{sec3} describes the models used in the study. Section \ref{sec4} presents and discusses the results. Section \ref{sec5} concludes this paper.

\section{Related Work} \label{sec2}

\subsection{Keyphrase Selection for Russian Texts}

Much of the current literature on keyphrase selection for Russian texts pays particular attention to unsupervised methods. Various studies \cite{guseva2024,khokhlova2022keyness,mitrofanova2022,morozov2023generation,sandul2018keyword} have assessed the efficacy of statistical methods, such as TFIDF, YAKE! \cite{campos2020yake}, RAKE \cite{rose2010automatic}, and graph-based methods, such as TextRank \cite{mihalcea2004textrank} and TopicRank~\cite{bougouin2013topicrank}. Some studies \cite{glazkova2024exploringfinetunedgenerativemodels,glazkova2023keyphrase,guseva2024,nguyen2021keyphrase,sokolova2017automatic} used the approaches based on machine learning, such as KEA \cite{witten1999kea} and pre-trained language models.

Researchers have identified the main challenges related to keyphrase selection. Morozov et al. \cite{morozov2023generation} identified the major difficulties related to keyphrase selection for Russian texts, including the need for lemmatization in the case of traditional unsupervised methods and a small number of existing datasets. The results obtained in \cite{guseva2024} showed a low coincidence rate between the keyphrases extracted using different methods (RAKE, RuTermExtract, KeyBERT, ChatGPT, etc.) and suggested that the choice of a  keyphrase selection method should be based not only on statistical criteria of the keyphrases, but also on their perception.

\subsection{Prompt-based Methods for Keyphrase Generation}

Recent advances in LLMs that can communicate with humans and generate coherent and meaningful responses \cite{achiam2023gpt,llama3modelcard,almazrouei2023falcon,jiang2023mistral,team2023gemini} are beneficial for developing effective solutions to various NLP tasks, including keyphrase selection. To date, several studies have investigated zero-shot and few-shot prompt-based methods for keyphrase generation. So far, most research on prompt-based keyphrase generation has focused on evaluating datasets in English.

Attempts have been made to evaluate the ability of prompt-based LLMs to generate keyphrases. Song et al. \cite{song2023large,song2023chatgpt} verified a zero-shot performance of ChatGPT on four keyphrase extraction datasets. In \cite{martinez2023chatgpt}, the performance of ChatGPT using an instructional prompt was compared with the results of several neural models for keyphrase selection. ChatGPT outperformed other models on all benchmarks, notably in handling long documents and non-scientific texts. The paper \cite{chataut2024comparative} explored keyphrase extraction using Llama2-7B, GPT-3.5, and Falcon-7B for two English scientific datasets and emphasized the role of prompt engineering in LLMs for a better keyphrase selection. The paper \cite{lee2023toward} examined the performance of Galactica, a model pre-trained on open-access scientific text and data \cite{taylor2022galactica}, for generating keyphrases. The authors of \cite{kang2023ai} compared author keyphrases from papers on the digital divide with those generated using BERT and ChatGPT. The correlation between author keyphrases and ChatGPT was higher than that between author keyphrases and BERT. Er et al. \cite{er2024llm} studied the performance of unsupervised methods such as TF-IDF, YAKE!, and TextRank against several prompt-based methods based on GPT3.5, GPT4, and Gemini and fine-tuned models, such as T5 \cite{raffel2020exploring} and BART \cite{lewis2020bart}. This study was performed on a Turkish corpus of customer reviews. The most accurate keyphrases were generated by GPT4 in a few-shot setting.
\newline

Overall, the review of related work reveals that prompt-based methods demonstrate high results for keyphrase generation. The scholars reported that LLMs generate keyphrases more accurately in a few-shot setting. Most current studies have been conducted on English-language datasets. Although LLMs have shown impressive multilingual performance, their ability to generate keyphrases for Russian texts has not been investigated. Current studies use various metrics to evaluate exact matches and semantic similarity between selected keyphrases and the gold standard keyphrases. Researchers \cite{alami2020automatic,guseva2024} emphasize that different aspects of keyphrase selection require different metrics including human evaluation.

\section{Methods} \label{sec3}

\subsection{Dataset}

The study used the Math\&CS dataset\footnote{\url{https://data.mendeley.com/datasets/dv3j9wc59v/1}} presented in \cite{morozov2023generation}, which consists of 8348 abstracts of research papers in the fields of mathematics and computer science sampled from the Cyberleninka online library with keyphrases tagged by the authors of the papers. Each dataset example includes an abstract and its corresponding list of keyphrases. Math\&CS contains keyphrases that are present as well as those that are absent in the abstract. The training set contains 5844 examples. 554 examples from the training set contain only present keyphrases, 655 examples include only absent keyphrases, other 4635 examples involve mixed keyphrases.

The dataset statistics is presented in Table \ref{table_data}. The average numbers of tokens and sentences are calculated using the NLTK package \cite{bird2006nltk}. The bottom row presents the overall percentage of absent keyphrases in the dataset. 

\begin{table}
\scriptsize
\centering
\caption{Dataset statistics}
\begin{tabular}{|l|l|}
\hline
Characteristic & Value \\ \hline
Train size & 5844  \\ \hline
Texts with present keyphrases & 554\\\hline
Texts with absent keyphrases & 655\\\hline \hline
Test size & 2504 \\ \hline \hline
Avg number of sentences & 3.73$\pm$2.75  \\ \hline
Avg number of tokens & 74.16$\pm$61.65  \\ \hline
Avg number of keyphrases per text & 4.34$\pm$1.5 \\ \hline
Absent keyphrases, \% & 53.66  \\\hline
\end{tabular}
\label{table_data}
\end{table}

\subsection{Models}

For prompt-based learning, we used three open-source instruction-following LLMs. To obtain more reliable results, we choose one English-oriented LLM and two models specifically adapted for the Russian language using different approaches.

\begin{itemize}
    \item Saiga/Mistral 7B (\textbf{Saiga})\footnote{\url{https://huggingface.co/IlyaGusev/saiga_mistral_7b_lora}} \cite{gusev_rulm_2023}, a Russian Mistral-based chatbot adapted by training LoRA adapters. This model was tuned on a dataset of ChatGPT-generated chats in Russian. 
    \item Vikhr-7B-instruct\_0.4 (\textbf{Vikhr})\footnote{\url{https://huggingface.co/Vikhrmodels/Vikhr-7B-instruct_0.4}} \cite{nikolich2024vikhr}. Contrary to Saiga, Vikhr uses adapted tokenizer vocabulary as well as continued pre-training and instruction tuning of all weights instead of LoRA adapters.
    \item Mistral-7B-Instruct-v0.2 (\textbf{Mistral})\footnote{\url{https://huggingface.co/mistralai/Mistral-7B-Instruct-v0.2}} \cite{jiang2023mistral}, an instruct fine-tuned version of Mistral-7B-v0.2, which is one of the most popular open-source LLMs. Although the Mistral's performance decreases for non-English languages \cite{marchisio2024understanding,ochieng2024beyond}, it still demonstrates meaningful results on various evaluation tasks for the Russian language \cite{fenogenova2024mera}.
\end{itemize}

These three models were prompted using two approaches: zero-shot and few-shot. Following \cite{er2024llm,kang2023ai}, we used a simple prompt text (Table \ref{table_prompts}) with special tokens in accordance with the requirements of the models used. The prompts were written in Russian. Three strategies for creating few-shot prompts were compared. In the first strategy (\textbf{random keyphrases}), three random examples from the training set were selected for each abstract from the test set. The second strategy (\textbf{present keyphrases}) included only the examples that contained present keyphrases (see Table~\ref{table_data}). The third strategy (\textbf{absent keyphrases}) focused on the examples with absent keyphrases. All instruction-following LLMs were required to generate a maximum of 100 tokens with a temperature of 0.5.

\begin{table}[t]
\scriptsize
\centering
\caption{Prompts}
\begin{tabular}{|p{5cm}|p{7cm}|}
\hline
\textbf{Zero-shot} & \textbf{Few-shot}  \\
\hline
\selectlanguage{russian}Сгенерируй ключевые слова для научной статьи по тексту аннотации. Ключевые слова выведи в одну строку через запятую.\newline
Текст аннотации: \{\selectlanguage{british}text\}
\newline
\textit{
\newline\selectlanguage{british}Generate keyphrases for a scientific paper using the given abstract. Keyphrases are written in one line and separated by commas.\newline
Abstract: \{\selectlanguage{british}text}\} & 
\selectlanguage{russian}Сгенерируй ключевые слова для научной статьи по тексту аннотации. \newline
Текст аннотации: \{\selectlanguage{british}text example 1\}\selectlanguage{russian}\newline
Ключевые слова: \{\selectlanguage{british}keyphrases example 1\}\selectlanguage{russian}\newline
Текст аннотации: \{\selectlanguage{british}text example 2\}\selectlanguage{russian}\newline
Ключевые слова: \{\selectlanguage{british}keyphrases example 2\}\selectlanguage{russian}\newline
Текст аннотации: \{\selectlanguage{british}text example 3\}\selectlanguage{russian}\newline
Ключевые слова: \{\selectlanguage{british}keyphrases example 3\}\selectlanguage{russian}\newline
Текст аннотации: \{\selectlanguage{british}text\}
\newline
\textit{
\newline\selectlanguage{british}Generate keyphrases for a scientific paper using the given abstract.\newline
Abstract: \{\selectlanguage{british}text example 1\}\selectlanguage{russian}\newline
Keyphrases: \{\selectlanguage{british}keyphrases example 1\}\selectlanguage{russian}\newline
Abstract: \{\selectlanguage{british}text example 2\}\selectlanguage{russian}\newline
Keyphrases: \{\selectlanguage{british}keyphrases example 2\}\selectlanguage{russian}\newline
Abstract: \{\selectlanguage{british}text example 3\}\selectlanguage{russian}\newline
Keyphrases: \{\selectlanguage{british}keyphrases example 3\}\selectlanguage{russian}\newline
Abstract: \{\selectlanguage{british}text}\}\\
\hline
\end{tabular}
\label{table_prompts}
\end{table}

\begin{itemize}
    \item \textbf{mT5}\footnote{\url{https://huggingface.co/google/mt5-base}} \cite{xue2021mt5}, a multilingual text-to-text transformer pre-trained on a Common Crawl-based dataset covering 101 languages. The architecture and training procedure are similar to T5 \cite{raffel2020exploring}.
    \item \textbf{mBART}\footnote{\url{https://huggingface.co/facebook/mbart-large-50}} \cite{tang2020multilingual}, a machine translation sequence-to-sequence model that uses the same baseline architecture as that of BART \cite{lewis2020bart}. mBART was trained on more than 50 languages with a combination of span masking and sentence shuffling.
    \item \textbf{YAKE!} \cite{campos2020yake}, a method that uses statistical features extracted from single documents to select the most relevant keyphrases of a text. We applied YAKE! for the Russian language using keyphrases.mca.nsu.ru\footnote{\url{https://keyphrases.mca.nsu.ru/}}.
    \item \textbf{RuTermExtract} \cite{rutermextract}, a term extraction tool for Russian based on statistical analysis, which also performs a rule-based lemmatization of keyphrases.
\end{itemize}

mT5 and mBART were fine-tuned on the training set for a keyphrase generation task for ten epochs with a maximum sequence length of 256 tokens and a learning rate of 4e-5. The number of keyphrases generated by the models for the abstracts from the test set was not limited. For RuTermExtract and YAKE!, the number of extracted keyphrases ($k$) was set to 5, 10, and 15. The results include the best scores for each metric.

\section{Experiments and Results}\label{sec4}

\subsection{Comparing Models in Terms of Evaluation Metrics}\label{subsec}

We used the following metrics for evaluation: the full-match F1-score, ROUGE-1 \cite{lin2004rouge}, and BERTScore \cite{zhangbertscore}. The full match F1-score assesses the number of exact matches between the list of keyphrases tagged by the author of the paper and the keyphrases selected by the model. To calculate F1-score, the keyphrases were lemmatized to reduce the number of mismatches. Next, the True Positive value was calculated as the size of the intersection between the sets of produced and author's keyphrases. The False Positive value represented the difference between the set of generated keyphrases and the author's ones. The False Negative value was defined as the difference between the sets of author's and generated keyphrases. ROUGE-1 accesses the number of matching unigrams between the selected keyphrases and the author's list of keyphrases. ROUGE-1 scores were calculated without preliminary lemmatization of keyphrases. BERTScore assesses the cosine similarity between the contextual embeddings of the tokens for both lists of keyphrases. The multilingual BERT (mBERT) \cite{devlin-etal-2019-bert} was used as a basic language model. For ROUGE-1 and BERTScore, the lists of keyphrases were represented as comma-separated strings. 

Table \ref{table_main_results} presents the results of the model comparison in terms of the selected metrics. The highest result for each metric in shown in bold, the second best result is underlined, and the third best result is double underlined. As expected, few-shot learning increased the performance of the prompt-based methods. Nevertheless, our results demonstrated that even the use of a zero-shot approach with a fairly simple prompt showed the performance comparable to other models. The best and second best scores in terms all metrics were obtained using Saiga\&few-shot (random keyphrases) and Saiga\&few-shot (present keyphrases). The third best result for different metrics was shown by mBART and Saiga\&few-shot (absent keyphrases). Among both fine-tuned models and unsupervised methods, the highest scores were achieved by mBART. mT5, YAKE!, and RuTermExtract performed worse than mBART in terms of all metrics. RuTermExtract demonstrated the highest scores with $k=10$. YAKE! showed the highest BERTScore using $k=5$ and the highest ROUGE-1 and F1-score using $k=15$. 

The use of present or random keyphrases show similar results. For Saiga, present keyphrases slightly rose the ROUGE-1 score (+0.03\%) in comparison with random keyphrases. For Vikhr, the performance increased in terms of all metrics (BERTScore -- +0.69\%, ROUGE-1 -- +0.59\%, F1-score -- +0.61\%). The scores of Mistral did not increase. The use of absent keyphrases led to a decrease in all metrics.

Summarizing the results obtained from the prompt-based models, it can be stated that the models adapted for the Russian language outperformed the English-oriented Mistral model. As expected, the performance of few-shot models was higher than that of zero-shot models. The best results were achieved by few-shot models using random or present keyphrases.

\begin{table}[]
\scriptsize
\centering
\caption{Results, \%}
\begin{tabular}{|l|l|l|l|}\hline
Model & BERTScore & ROUGE-1 & F1-score \\\hline
Saiga\&zero-shot & 77.72 & 18.86 & 15.95 \\
Vikhr\&zero-shot& 76.16 & 18.13 & 14.46 \\
Mistral\&zero-shot & 74.27 & 13.77 & 10.82 \\\hline
Saiga\&few-shot (random keyphrases)& \textbf{79.5} & \underline{22.37} & \textbf{20.16} \\
Vikhr\&few-shot (random keyphrases)& 77.48 & 19.62 & 15.18 \\
Mistral\&few-shot (random keyphrases)& 74.85 & 16.3 & 15.08 \\\hline
Saiga\&few-shot (present keyphrases) & \underline{79.06} & \textbf{22.4} & \underline{19.34} \\
Vikhr\&few-shot (present keyphrases) & 78.17&20.21&15.79\\
Mistral\&few-shot (present keyphrases) & 73.73 & 15.11 & 14.52\\\hline
Saiga\&few-shot (absent keyphrases) & 78.5 & \underline{\underline{20.82}} & \underline{\underline{16.98}} \\
Vikhr\&few-shot (absent keyphrases) & 77.41& 18.65& 14.39\\
Mistral\&few-shot (absent keyphrases)& 72.17& 12.64& 11.57\\\hline\hline
mT5 & 76.07 & 15.14 & 13.41 \\
mBART & \underline{\underline{78.66}} & 19.26 & 16.84\\\hline
YAKE! & 69.13 & 6.47 & 6.03 \\
RuTermExtract & 75.95 & 15.12 & 11.02 \\\hline
\end{tabular}
\label{table_main_results}
\end{table}

\subsection{Human Evaluation}

\begin{figure}[h]
\begin{minipage}[h]{0.45\linewidth}
\center{\includegraphics[width=1\linewidth]{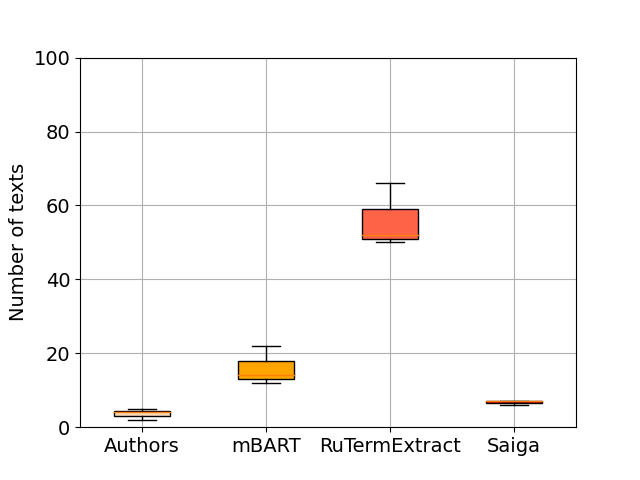}} a) \\
\end{minipage}
\hfill
\begin{minipage}[h]{0.45\linewidth}
\center{\includegraphics[width=1\linewidth]{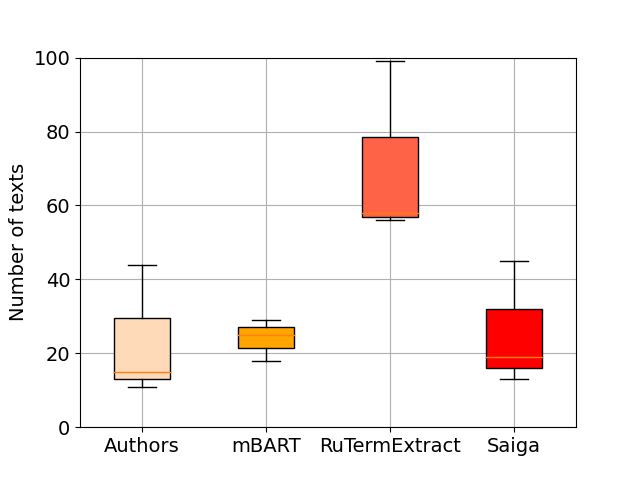}} \\b)
\end{minipage}
\vfill
\begin{minipage}[h]{0.45\linewidth}
\center{\includegraphics[width=1\linewidth]{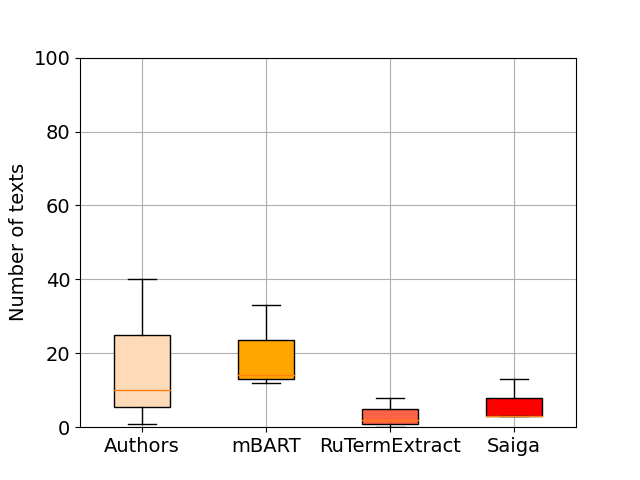}} c) \\
\end{minipage}
\hfill
\begin{minipage}[h]{0.45\linewidth}
\center{\includegraphics[width=1\linewidth]{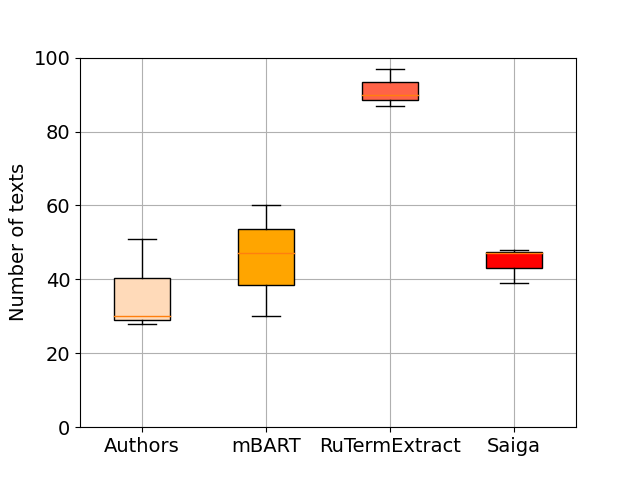}} d) \\
\end{minipage}
\caption{Human evaluation results: a) presence of grammar and spelling mistakes, b)
redundancy, c) insufficiency , d) presence of generic words.}
\label{ris:humanevaluation}
\end{figure}

In addition to calculating metrics, human evaluation was used to assess the quality of keyphrases.
To perform human evaluation, we selected Saiga\&few-shot (random keyphrases) (hereinafter -- Saiga), mBART, RuTermExtract as they demonstrated the highest performance among prompt-based methods, fine-tuned models, and unsupervised methods respectively. One hundred random texts from the test set were randomly selected. For each text, the outputs of Saiga, mBART, and RuTermExtract were collected (300 outputs in total). Then, three experts with a background of writing academic papers in computer science marked each output according to the following criteria: (a) \textbf{Presence of grammar and spelling mistakes}. True if the list of keyphrases contains any grammar or spelling mistakes; (b) \textbf{Redundancy}. True if keyphrases are redundant, for example, contain a lot of cognate words or synonyms; (c) \textbf{Insufficiency}. True if the list of keyphrases does not describe the content of the text well enough; (d) \textbf{Presence of generic words}. True if keyphrases contain generic words or phrases that do not describe the subject area, for example, ''paper'', ''study'', etc. In addition, the experts were asked to mark the authors' lists of keyphrases for the selected texts. The expert assessments are averaged and visualized in Figure~\ref{ris:humanevaluation}.

The results allow analysis of both model drawbacks and dataset limitations. First, the results reveal significant differences between the experts' assessments. As expected, there is a slight difference in the presence of grammar and spelling mistakes, since it is the most objective criterion. Other criteria, particularly redundancy, have more diverse assessments. A significant difference in expert assessments implies that the criteria for high-quality keyphrases are poorly formalized. Similar to other text generation tasks \cite{bhandari2020re,novikova2017we}, the choice
of metrics is influenced by a variety of factors, including the specific tasks, datasets, and application scenarios. Second, human evaluation indicates main strengths and weaknesses of each method. Thus, RuTermExtract extracts a large number of redundant lists of keyphrases and generic words. The keyphrases extracted by RuTermExtract contain the largest number of grammar and spelling mistakes in connection with rule-based lemmatization. However, the outputs of RuTermExtract are characterized by low insufficiency. mBART and Saiga are similar in redundancy and the number of generic words, while the number of spelling mistakes and insufficiency are lower for Saiga. The authors' keyphrases are the best in terms of the number of grammar and spelling mistakes and generic words and redundancy, but regarding insufficiency they are worse than RuTermExtract and Saiga. The results also reveal that the authors' keyphrases are not always optimal from the experts' perspective. Therefore, the selection of texts for training datasets or few-shot examples requires an additional expert assessment.

\subsection{Discussion and Limitations}

As in previous research carried out on text corpora in other languages \cite{er2024llm,martinez2023chatgpt,song2023chatgpt}, prompt-based LLMs have shown promising results in keyphrase generation for Russian. These models offer several advantages, including the capacity for effective few-shot learning, as well as the lemmatization of generated keyphrases. We have tested open-source instruction-following LLMs on scientific texts and demonstrated that prompt-based methods have the ability to generate quite informative and comprehensive keyphrases according to human evaluation results. Additionally, our findings suggest that all the considered prompt-based LLMs perform better using the examples with present keyphrases or random examples from the dataset than using the examples with absent keyphrases. 

The current study is limited by the dataset features. First, we believe that prompt-based LLMs have a greater potential for keyphrase generation in other domains and text genres, namely, for news. However, domain efficiency and transferability needs further research. The effectiveness of keyphrase generation using prompt-based LLMs on long scientific texts, such as full paper texts, also requires further investigation. Second, since the examples for fine-tuning and creating prompts were obtained from the dataset, some dataset features can be reflected in generated keyphrases. This study is also limited by the use of a simple prompt structure. Various prompt formulations can be explored in further research.

\section{Conclusion}\label{sec5}

This study explored the ability of prompt-based LLMs to generate keyphrases for Russian scientific abstracts. We compared prompt-based methods with fine-tuned models and unsupervised methods and found that prompt-based LLMs achieve superior performance in comparison with baselines, even when employing basic text prompts. We employed different strategies for selecting keyphrase examples in a few-shot setting and observed that the use of the examples containing only absent keyphrases leads to lower performance. Finally, we provided the results of human evaluation across three models and discussed their strengths and limitations.

\section*{Acknowledgements}\label{sec6}

We are grateful to Nadezhda Zhuravleva (Center for Academic Writing ''Impulse'', University of Tyumen) for her assistance with the English language.

%
%
%
\bibliographystyle{splncs04}
\bibliography{aist}
\end{document}